\documentclass{egpubl}
\usepackage{pg2026}

\SpecialIssuePaper         

\CGFStandardLicense

\usepackage[T1]{fontenc}
\usepackage{dfadobe}  

\usepackage{cite}  
\BibtexOrBiblatex
\electronicVersion
\PrintedOrElectronic
\usepackage{graphicx}
\ifpdf \pdfcompresslevel=9 \fi

\usepackage{egweblnk} 
\usepackage{tikz-cd}

\newcommand{\Eq}[1]  {Eq.~(\ref{eq:#1})}

\newcommand{\Fig}[1] {Fig.~\ref{fig:#1}}

\newcommand{\Tbl}[1]  {Table~\ref{tbl:#1}}

\newcommand{\Sec}[1] {Section~\ref{sec:#1}}

\usepackage{graphicx}
\usepackage{fmtcount}
\usepackage{epsfig} 
\usepackage{amsmath}

\usepackage{amssymb}  
\usepackage{booktabs} 
\usepackage{subcaption} 
\usepackage{lipsum}
\usepackage{arydshln} 
\usepackage{comment}
\usepackage{algorithm2e}
\usepackage{algpseudocode} 
\usepackage{bm}
\usepackage[cjk]{kotex}
\usepackage{cuted} 
\usepackage{caption}
\usepackage{xcolor}
\usepackage{color, soul}
\usepackage{multirow}
\usepackage[table]{xcolor}
\SetKwInput{KwIn}{Input}
\SetKwInput{KwOut}{Output}
\usepackage[table]{xcolor}
\usepackage{booktabs}
\SetAlgoNoLine         
\DontPrintSemicolon    
\usepackage{arydshln}

\usepackage{tikz}

\title{Camera Splatting for Continuous View Optimization}

\author[G.\ Lee et al.]
{\parbox{\textwidth}{\centering Gahye Lee\orcid{0009-0006-3573-4954}
        Hyomin Kim\orcid{0000-0002-2162-4627}
        Gwangjin Joo\orcid{0000-0003-1006-8351}
        Jooeun Son\orcid{0009-0003-6916-9912}
        Hyejeong Yoon\orcid{0009-0004-4423-8544}
        Seungyong Lee\thanks{Corresponding author.}\orcid{0000-0002-8159-4271}
}
        \\
{\parbox{\textwidth}{\centering POSTECH, South Korea
       }
}
}

\begin{document}

\teaser{
\includegraphics[width=0.9\textwidth,trim=0.4cm 7.cm 0.3cm 0.8cm,clip]{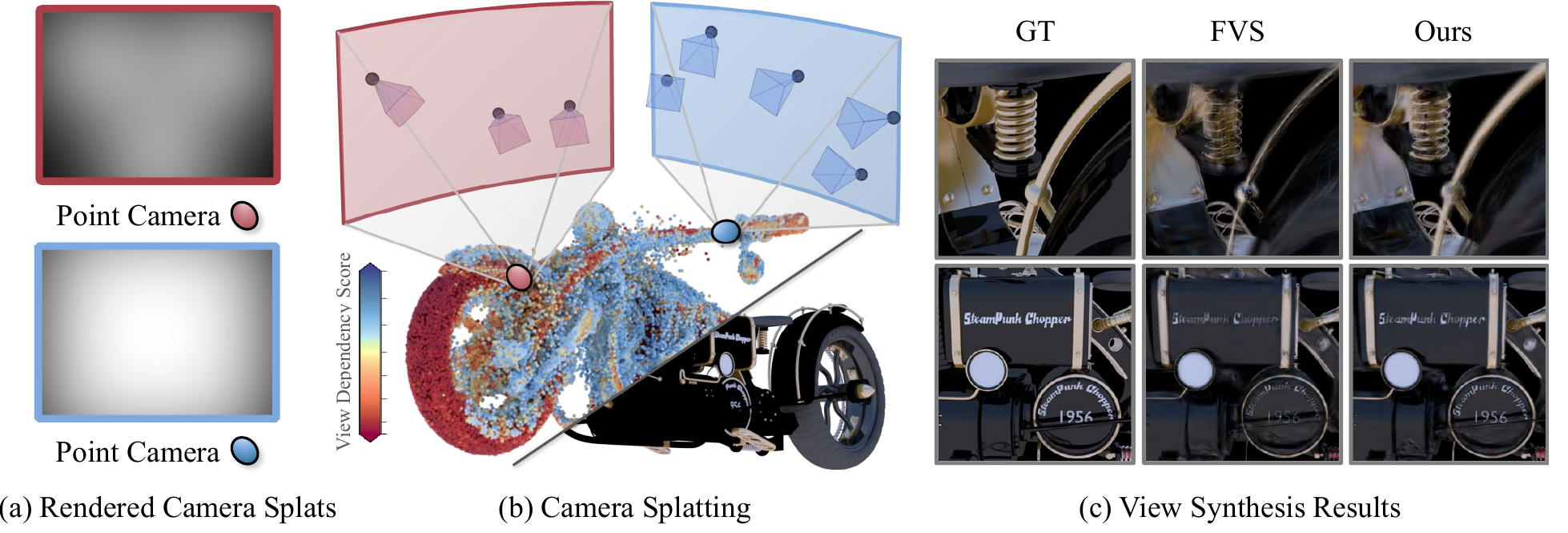}
 \centering
  \caption{
We propose {\em Camera Splatting}, a novel view optimization framework for radiance field reconstruction. 
Each camera is modeled as a 3D Gaussian, referred to as a {\em camera splat}, and virtual cameras, termed {\em point cameras}, are placed at 3D points sampled near the surface to observe the distribution of camera splats. 
View optimization is achieved by {\em continuously and differentiably} refining the camera splats through rendering from the point cameras, in a manner similar to the original 3D Gaussian splatting, so that resulting cameras are effectively distributed. 
Compared to the Farthest View Sampling (FVS) approach,   
which shows competitive performance in NeRF Director~\cite{xiao2024nerf},
our optimized views demonstrate superior performance in capturing complex view-dependent phenomena, including intense metallic reflections and intricate textures such as text.} 
\label{fig:teaser}
}

\maketitle

\begin{abstract}

Given limited camera budgets, identifying informative views is essential for effectively reconstructing a radiance field to achieve high-quality novel view synthesis.  
These views should adaptively sample the radiance field based on the geometry and view-dependent appearance of the scene. 
Existing methods typically optimize views by selecting from a predefined camera set or adopting a greedy next-best-view strategy, limiting the flexibility and accuracy of the optimization.
In this paper, 
we propose a novel continuous view optimization framework, called {\em Camera Splatting}, which enables joint optimization of multiple views. 
Our key idea is to represent each camera as a specialized 3D Gaussian, called {\em camera splat}, allowing for gradient-based optimization within a continuous camera parameter space. 
To evaluate view optimality, we introduce virtual cameras, called {\em point cameras} 
placed at sampled points in the scene,  
which render camera splats and provide feedback on their spatial positions and orientations. 
The resulting optimized views effectively capture the view-dependent appearance of the scene at the sampled points. 
Integrated into a Gaussian splatting pipeline, our approach can efficiently and simultaneously optimize a large number of views. 
Quantitative and qualitative evaluations demonstrate the effectiveness of the views optimized by Camera Splatting for novel view synthesis under a limited camera budget.

\ccsdesc[500]{Computing methodologies~Image-based rendering}
\ccsdesc[300]{Computing methodologies~Reconstruction} 
\printccsdesc   
\end{abstract}  

\vspace*{-0.44cm}
\section{Introduction}
\label{sec:intro}  
Novel view synthesis aims to generate unseen views of a  scene from a set of captured images. 
Central to this task is the radiance field, a continuous representation encoding scene geometry and view-dependent appearance~\cite{mildenhall2020nerf, barron2022mipnerf360, fridovich2022plenoxels, kerbl20233d}. 
Formally, a radiance field is defined as a continuous mapping from a 5D input, 
consisting of 3D spatial coordinates $x$ and 2D directions $\omega$, to color $c$ and density $\sigma$.

Reconstructing a radiance field requires sampling across both $x$ and $\omega$.
Such sampling is instantiated by the rays emitted from the cameras, which sample the radiance field along different spatial locations and directions.
A dense, uniform capture tends to improve reconstruction quality,
but constraints such as a limited camera budget and computational resources make this impractical.
Therefore, selecting the most informative views is essential for efficiently reconstructing the radiance field, enabling high-quality novel view synthesis under a limited camera budget~\cite{kopanas2023improving, li2024frequency,jiang2023fisherrf, pan2022activenerf, lyu2024manifold}.

With recent advances, view optimization methods prioritize improving the quality of novel view synthesis. 
Several methods~\cite{chen2024gennbv, zhan2022activermap, jiang2023fisherrf,pan2022activenerf} measure the uncertainty of the reconstructed radiance field to identify regions that require additional sampling. 
However, they typically rely on sequential view selection, where views are selected one at a time, rather than jointly optimizing all views. This can fail to consider dependencies among selected views and lead to suboptimal solutions.
Other methods select views from a predefined discrete set of candidate cameras~\cite{xiao2024nerf,pan2022activenerf, jiang2023fisherrf, kopanas2023improving}, which limits the search space and excludes potentially more informative viewpoints. 

To address these limitations, view optimization should be performed in a continuous camera parameter space rather than a predefined discrete candidate set. This enables the search for informative viewpoints without restricting the optimizer to a fixed camera pool. Moreover, all views should be optimized jointly instead of being selected sequentially, allowing inter-view dependencies to be considered during optimization. This strategy should also be computationally efficient and scalable to large camera budgets.

In this paper, we present {\em Camera Splatting}, a novel view optimization framework inspired by 3D Gaussian Splatting (3DGS)~\cite{kerbl20233d}, which supports continuous and joint optimization of multiple cameras.
Our core innovation is to represent each camera as a specialized 3D Gaussian, called a \emph{camera splat}, that provides a continuous representation of camera parameters.
We further introduce virtual cameras, termed \emph{point cameras}, to evaluate the quality of radiance field sampling achieved by the camera splats. 
Leveraging the differentiable 3DGS optimization pipeline, our framework directly optimizes camera parameters by rendering camera splats from point cameras. 
This enables efficient and scalable optimization that can handle a large number of cameras simultaneously. 

To adapt Camera Splatting to a target scene, our framework first builds a proxy geometry from sparse input views.
The proxy geometry serves as an approximate spatial sampling region that guides where to place point cameras.
Furthermore, we compute the appearance variation across the input views to estimate the directional sampling density, which is incorporated into the Camera Splatting objective. 
Our framework then jointly optimizes the parameters of the camera splats to identify informative views.

We validate Camera Splatting on both object-scale and scene-scale datasets, including challenging scenes with complex materials and detailed textures. Across varying camera budgets, our method achieves faster optimization and higher reconstruction quality than existing approaches.

In summary, our contributions are as follows:
\begin{itemize}
    \item We propose {\em Camera Splatting}, a continuous view optimization framework that enables the joint optimization of multiple cameras for high-quality radiance field reconstruction.
    \item We introduce a {\em camera splat}, a Gaussian-based representation of a camera to be optimized, and a {\em point camera}, a virtual camera model for evaluating the radiance field sampling induced by the camera configuration.
    \item We demonstrate that Camera Splatting achieves faster optimization and higher reconstruction quality than existing view optimization approaches across both object-scale and scene-scale datasets.
\end{itemize}
\section{Related Work}
\label{sec:relwork}

\paragraph*{View optimization for geometry reconstruction}
The main goal of view optimization methods for geometry reconstruction is to maximize coverage to ensure sufficient geometric constraints between views~ \cite{  zhou2020offsite, mendoza2020supervised, fan2016automated, scott2003view,krainin2011autonomous}.
Aerial Path Planning (APP)~\cite{smith2018aerial} introduces reconstructability heuristics for stereo matching from pairwise relationships among views.
While APP leverages the downhill simplex method for optimization, Offsite Aerial Path Planning (OAPP)~\cite{zhou2020offsite} employs bundle adjustment. 
Other methods such as ~\cite{roberts2017submodular} propose local surface coverage through angular diversity and select the best trajectories to maximize global scene coverage.
Collectively, these methods define view quality mainly through geometric coverage and reconstructability.

\paragraph*{View optimization for novel view synthesis}
Recent work on view optimization for radiance field reconstruction has primarily measured view quality using uncertainty-, information-based criteria defined on the current reconstruction state~\cite{  jin2023neu, kendall2017uncertainties, sunderhauf2023density, lee2022uncertainty, savant2024modeling,  ran2023neurar}. 
FisherRF~\cite{jiang2023fisherrf} leverages Fisher Information derived from the Hessian matrix of the loss function to select views that maximize information gain.
ActiveNeRF~\cite{pan2022activenerf} defines uncertainty as color variance estimated from each spatial position and trains the variance using the NeRF~\cite{mildenhall2020nerf} framework. 
Manifold sampling~\cite{lyu2024manifold} models the radiance field using stochastic variables, considering the sampling variance as an uncertainty measure. 
ACP~\cite{kopanas2023improving} explicitly incorporates observation frequency and angular uniformity to balance view frequency and directional coverage. These methods assess view quality through uncertainty, information gain, or observation statistics on the radiance field, whereas our framework evaluates view quality through the radiance field sampling induced by the camera configuration.


\paragraph*{Optimization strategies} 
Different strategies have been adopted to identify optimal views based on various definitions of view optimality.
For computational efficiency and simplicity, most existing methods rely on discrete optimization, selecting views from 
a fixed set of candidate cameras~\cite{yi2023render, pan2022activenerf, jiang2023fisherrf, kopanas2023improving, zhou2020offsite, beder2006determining, dunn2009next}. 
The final solution is constrained by the coverage of the candidate view set. Although increasing the number of candidates enlarges the feasible solution space, it also raises the combinatorial complexity of multi-view selection, making it necessary in practice to optimize over a limited candidate set.
To alleviate this, some works~\cite{roberts2017submodular} optimize rotation and translation parameters separately, simplifying the search at the cost of restricting joint updates in the camera pose space.

Some methods instead formulate view optimization in a continuous camera parameter space.
APP~\cite{smith2018aerial} incrementally optimizes the camera configuration by 
updating one view at a time in the continuous parameter space.
Manifold Sampling~\cite{lyu2024manifold} proposes a differentiable uncertainty metric for radiance fields, enabling gradient-based optimization of individual views.

Although various strategies have been explored for view optimization, most existing methods optimize camera configurations sequentially, adding one view at each step based on the current partial configuration. Such incremental formulations can be susceptible to local minima. In contrast, our framework performs joint optimization over multiple camera views in a continuous parameter space. Leveraging a differentiable rendering inspired by 3D Gaussian Splatting~\cite{kerbl20233d}, our method enables efficient and scalable optimization over the full camera set.


\section{Preliminaries}
\label{sec:method}


\subsection{Radiance Fields}
A 3D scene can be represented as a \emph{radiance field}, which models scene appearance as a function of spatial position and viewing direction. Radiance fields provide a unified representation of geometry and appearance, making them a standard formulation for 3D scene reconstruction and novel view synthesis.
Formally, a radiance field $F$ is defined as
\begin{equation}
F : (\mathbf{x}, \omega) \mapsto (\sigma, \mathbf{c}),
\end{equation}
where $\mathbf{x} \in \mathbb{R}^3$ is a spatial position, $\omega \in \mathbb{S}^2$ is a viewing direction, and the output consists of a volumetric density $\sigma \in \mathbb{R}$ and a view-dependent color $\mathbf{c} \in \mathbb{R}^3$.

Given this formulation, radiance field reconstruction can be interpreted as a sampling problem over the joint domain of spatial positions and viewing directions. In practice, such sampling is instantiated by the image rays emitted from the cameras, which sample the radiance field along different spatial locations and directions. The fidelity of the reconstructed radiance field therefore depends on the sampling quality induced by the camera configuration. Under a limited camera budget, our goal is to identify a camera configuration whose rays effectively sample the target radiance field to support accurate reconstruction.

\subsection{3D Gaussian Splatting}
Our method builds on the parameterization and differentiable rendering mechanism of 3D Gaussian Splatting (3DGS)~\cite{kerbl20233d}. 3DGS represents a 3D scene with Gaussian primitives for efficient reconstruction and rendering.  
Each Gaussian is parameterized by a position \(\boldsymbol{\mu}_i\), a covariance matrix \(\Sigma_i\), an opacity \(\alpha_i\), and a color \(\mathbf{c}_i\). 
A set of 3D Gaussian primitives is rendered by projecting them onto the image plane and compositing their contributions using alpha blending.
The resulting rendering approximates the volume rendering process~\cite{kajiya1986rendering}:
\begin{equation}
L_o(r) \;=\; \sum_{i=1}^{N}\; T_i\,\alpha_i \,\mathbf{c}_i, 
\;\;\;\;\;
T_{i}
= \prod^{i-1}_{j=1}{(1-\alpha_{j})},
\label{eq:gs_render}
\end{equation}
where \(L_o(r)\) denotes the radiance along ray \(r\), and \(T_i\) denotes the accumulated transmittance. This rendering process is differentiable with respect to the Gaussian parameters.
Based on the differentiable 3DGS formulation, our method uses Gaussian primitives to continuously model cameras and optimize them through differentiable rendering. 
\section{Camera Splatting Framework}
\label{sec:cs} 
In the base formulation of \emph{Camera Splatting}, we represent the cameras to be optimized as \emph{camera splats}, and place virtual probes, termed \emph{point cameras}, at sampled spatial locations. Each point camera renders the camera splats to evaluate the radiance field sampling achieved under the current camera configuration.


 
\subsection{Camera Splat}
\label{sec:camera splat}  
We represent each camera as a specialized isotropic 3D Gaussian, called a \emph{camera splat}. 
The camera extrinsics are encoded directly as Gaussian properties, where the mean $\boldsymbol{u}_i$ is the camera center and the orientation $\boldsymbol{r}_i$ is aligned with the viewing direction. The camera intrinsics are encoded by a field of view (FoV) $\theta$ and assumed known, where $\theta$  defines the camera's view frustum. For simplicity, we share  $\theta$ across all camera splats, though each splat may instead take its own $\theta_i$ to represent cameras with different fields of view. The i-th camera splat is then defined as
\begin{equation}
C_i = \{ \boldsymbol{u}_i, \boldsymbol{r}_i, s, \theta, \alpha \}. 
\label{eq:my_equation}
\end{equation}
The scale $s$ is a global optimizable parameter shared across all camera splats, which indicates the extent of each Gaussian and thereby controls the spacing maintained between camera splats. The opacity $\alpha$ is a fixed value shared across all camera splats, so that a rendered image reflects the density of visible camera splats. The intuition behind this design is discussed in \Sec{base}.


\subsection{Point Camera}
To assess the radiance sampling quality induced by the camera splats, we introduce a virtual omnidirectional camera, termed the \emph{point camera}.
Unlike the camera splats, point cameras are not optimization variables. 
Instead, they are placed at sampled spatial positions that define where sampling density is evaluated. 
At each position, a point camera renders the camera splats into an omnidirectional image, where a higher intensity in a direction indicates that more camera splats are concentrated around it. This provides local supervision for optimizing the camera splats. Detailed point camera setup is provided in the supplementary material.



\begin{figure}
    \centering
    \includegraphics[width=0.8\linewidth,trim=0cm 9.cm 18.5cm 0cm, clip]{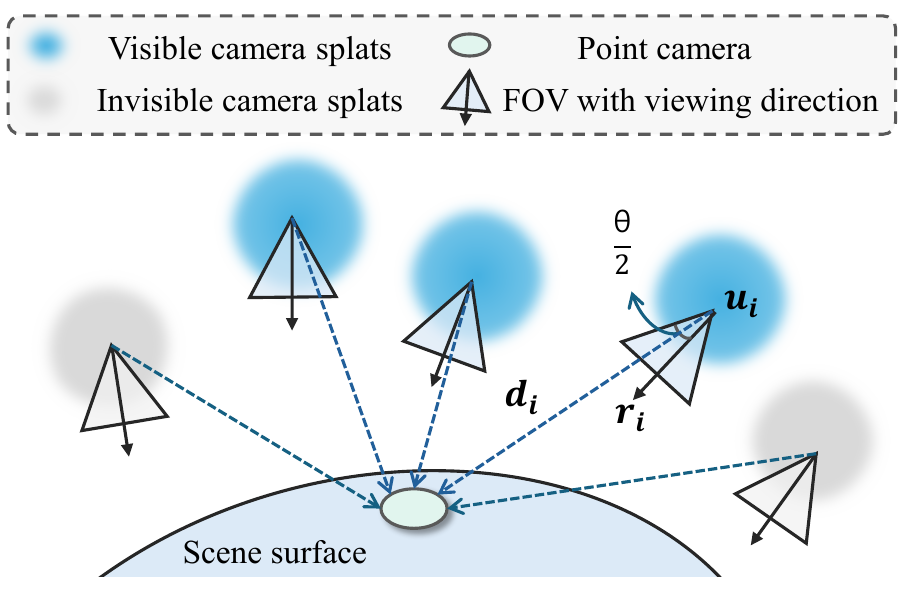}
    \caption{ Illustration of the camera splat visibility from a point camera. Blue splats satisfy the angular visibility condition $\angle(\boldsymbol{d}_i, \boldsymbol{r}_i) \leq \theta/2$, while grey splats fall outside this range and are treated as invisible.
        }
    \label{fig:base_render}
\end{figure}

\subsection{Camera Splat Rendering at Point Camera}  
\label{sec:rendering} 

To evaluate the current sampling density of the radiance field, we render the opacity values of camera splats from point cameras following the 3DGS rendering equation in \Eq{gs_render}.
During rendering, a camera splat is considered visible from a point camera if the angle between the splat's orientation $\boldsymbol{r}$ and the vector $\boldsymbol{d}$ from the splat to the point camera is within $\frac{\theta}{2}$, the half-angle of the field of view cone centered on $\boldsymbol{r}$ (\Fig {base_render}). 

Our rendering equation for a ray $\boldsymbol{\omega}$ from a point camera $P$ is
\begin{align}
    I_{render}(\boldsymbol{\omega}) = \sum_{i=1}^{N} T_i\,\alpha_i\,m_i, \qquad
    T_i = \prod_{j=1}^{i-1}(1-\alpha_j m_j),
    \\
    \text{where} \,\,\,\,
    m_i =
    \begin{cases}
    1 & \text{if } \angle(\boldsymbol{d_i}, \boldsymbol{r_i}) \leq \frac{\theta}{2} \\
    0 & \text{otherwise}
    \end{cases}, \notag
    \label{eq:complex_equation_1}
\end{align}
\noindent
where $N$ is the number of camera splats along the ray and $m_i$ is the visibility mask of the $i$-th splat.
The mask $m_i$ serves as a visibility gate that modulates the contribution of camera splats along each ray and the gradients are propagated to the parameters of splats with nonzero contribution.
Since all camera splats share a fixed opacity $\alpha$, a higher rendered intensity indicates more overlapping camera splats and thus denser sampling along that direction, while zero intensity means the direction is not sampled.

\subsection{Base Objective for Camera Splat Optimization}
\label{sec:base}
To optimize camera splats, we first construct a ground truth image that represents the ideal directional distribution of camera splats.
We define the ideal placement of camera splats as being uniformly distributed over all viewing directions from the point camera.  
For each point camera, we therefore define a base ground truth image $I_{\text{base}}$ as a constant monochromatic image whose intensity is set to the opacity $\alpha$. 
This  image encodes the desired uniform sampling of the radiance field over viewing directions at the point camera position.

Given $N$ point cameras, we define the base loss using the ground truth image $I_{\text{base}}$ and the rendered camera splat image $I_{\text{render}}$ as
\begin{equation}
\label{eq:base}
L_{\text{base}}=\frac{1}{N}\sum_{i=1}^{N}\left\| I_{\text{render}}^{(i)} - I_{\text{base}}^{(i)}\right\|_2^2.
\end{equation}
This loss encourages a uniform directional distribution of camera splats around each point camera. Since the projected splats have identical opacity and isotropic scale, matching a monochromatic target image requires them to be distributed evenly over viewing directions rather than concentrated in a particular region. This results in regular spacing with mild overlap among neighboring splats, thereby approximating uniform directional sampling.

\paragraph*{Shared scale $s$ and fixed opacity $\alpha$}
We use a shared scale parameter $s$ and a fixed opacity value $\alpha$ for all camera splats to regularize the optimization under \Eq{base} and avoid degenerate solutions. 
With per-splat scale, a small number of splats could expand and match the target image without forming a uniform distribution. 
Likewise, with per-splat opacity, several splats could cluster in a local region and match the target through their accumulated intensity. 
Such configurations satisfy the objective but fail to produce the intended regular directional distribution. 
By sharing $s$ and fixing $\alpha$, we suppress such degenerate solutions and encourage a more uniform directional distribution of camera splats.



\section{Continuous View Optimization} 
\label{sec:opt}  
\Sec{cs} introduced the base Camera Splatting framework, comprising the camera splat representation and the base objective. To adapt this framework to a target scene, we construct radiance field sampling targets from sparse input images in both spatial and angular domains. The spatial sampling target represents the scene’s surface support, indicating where samples should be concentrated in 3D space. The angular target represents a spatially varying sampling density, indicating how finely viewing directions should be sampled to capture view-dependent appearance. These targets are incorporated into the Camera Splatting objective for scene-specific view optimization.

\subsection{Spatial and Directional Sampling }  
\label{sec:targetsampling}

\paragraph*{Proxy geometry for spatial sampling} 
\label{sec:proxy} 
Effective radiance field reconstruction requires spatial sampling near the scene surface. We use a proxy geometry to approximate the surface and place point cameras around it, ensuring sufficient coverage of the scene surface during view optimization.

Since our method requires only a coarse approximation of the scene surface, the proxy geometry need not capture fine geometric details. This makes the framework practical, as it can be obtained from a small number of RGB images using recent feed-forward reconstruction methods such as DUSt3R~\cite{Wang_2024_CVPR_dust3r} and VGGT~\cite{wang2025vggt}. In our implementation, we use the point cloud reconstructed by VGGT. The effect of proxy geometry quality on view optimization is analyzed in our experiments.


\begin{figure}
    \centering
    \includegraphics[width=1\linewidth]{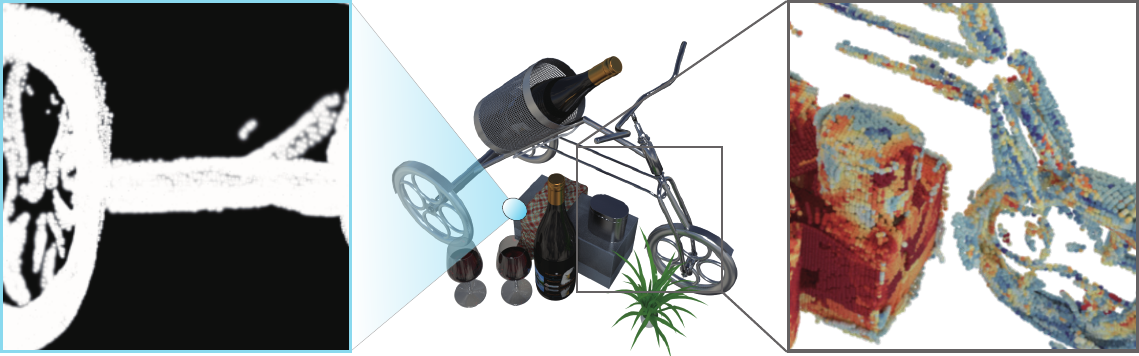}
    \caption{
        Visualization of the self-occlusion mask (left) and the estimated View-Dependency Score (VDS) (right). In the VDS map, blue indicates higher directional sampling density and red indicates lower density; metallic regions such as the wheel exhibit higher scores.
        }
    \label{fig:occlusion}
\end{figure}

\paragraph*{View Dependency Score for directional sampling}
\label{sec:vdsf}
Different materials exhibit distinct reflectance properties, leading to spatially varying degrees of view-dependent appearance. Accordingly, for faithful radiance field reconstruction, the directional sampling density should vary across surface locations, with locations exhibiting stronger view dependency sampled more densely.
We quantify this requirement using a View Dependency Score (VDS), defined for each surface point as its relative directional sampling weight.

To estimate this score, we introduce a View Dependency Score Function (VDSF), which maps the observed intensity variance $v$ at a surface point $p$ to the view dependency score $s$:
\begin{equation}
    s = \mathrm{VDSF}(v).
    \label{eq:vdsf}
\end{equation}
For each surface point $p$, we compute $v$ as the variance of the intensities of $p$ across the sparse input views in which it is visible.
We use $v$ as a proxy for local view dependency, since points with stronger view-dependent appearance tend to exhibit larger intensity variation across views.

The resulting score $s$ is min-max normalized across all surface points and shifted so that the minimum is $1$, ensuring that every point receives a baseline directional sampling density while more view-dependent points receive proportionally more.
This score is incorporated into the construction of the ground truth image for each point camera in \Eq{image_intensity}.

In practice, we parameterize VDSF as a cubic polynomial fitted to synthetic data spanning diverse materials and lighting conditions.
We choose this form empirically, as it is expressive enough to capture the variance-to-density relationship
without introducing unnecessary model complexity.
Additional details on data generation, model coefficients, and design ablations are provided in the supplementary material.



 
\begin{figure*}[h!]
    \centering 
    \includegraphics[width=1\textwidth,trim=0cm 6.5cm 0cm 7.3cm, clip]{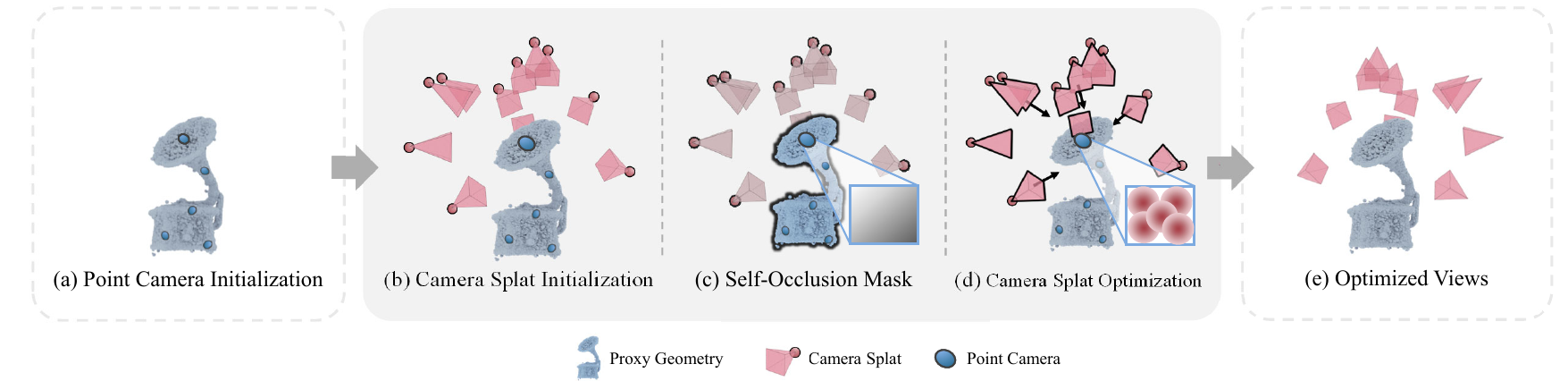} 
    \caption{
        Overall process of view optimization using our Camera Splatting framework,
        highlighting the initialization, core components of the framework, and the resulting optimized view configuration.  
        } 
    \label{fig:process}
    \vspace{-0.3cm}
\end{figure*}

\subsection{Objective Function}   
To incorporate the spatial and directional sampling targets into the optimization, we extend the base objective in \Eq{base} with a VDS-based image loss. Additionally, we introduce two regularization terms that enforce valid camera configurations.
\paragraph*{VDS-based image loss} 
We control the desired sampling density at each point camera $P_i$ by scaling the base ground truth image $I_{base}$ using the estimated view-dependency score: 
\begin{equation}
    I_{gt}(P_i) = \text{VDSF}(v_{P_i}) \times I_{base}.
    \label{eq:image_intensity}
\end{equation} 
where $v_{P_i}$ is the observed intensity variance at the spatial location of point camera $P_i$, computed from the sparse input views. Examples of the estimated VDS are illustrated in \Fig{occlusion}, right.  

This VDS-based scaling strategy adjusts the directional sampling density at each point camera \(P_i\). Specifically, a larger \(I_{gt}\) corresponds to a higher target intensity at \(P_i\), which requires more overlap among neighboring camera splats. This increased overlap results in denser directional sampling.

Given $N$ point cameras and camera splats $C$, we define the image loss $\mathcal{L}_{\text{image}}$ as the Mean Squared Error (MSE) between the rendered camera splat images $I_{\text{render}}$ and the ground truth images $I_{\text{gt}}$, with 
self-occlusion mask $I_{occ}$:
\begin{equation}
    \mathcal{L}_{\text{image}}
    = \frac{1}{N}\sum_{i=1}^{N} \left\| \bigl(I_{\text{render}}^{(i)} - I_{\text{gt}}^{(i)}\bigr) \odot (1-I_{\text{occ}}^{(i)}) \right\|_2^2,
    \label{eq:image_loss}
\end{equation}
where $\odot$ denotes element-wise multiplication.
The self-occlusion mask $I_{occ}^{(i)}$ is obtained by rendering the proxy geometry from each point camera $P_i$ as shown in \Fig{occlusion}, left. It suppresses contributions from camera splats that are occluded along the rays of  $P_i$, so that each camera splat receives gradients only from point cameras from which it is visible. 




\paragraph*{Directional regularizer}
To encourage camera splats to orient toward under-covered surface regions, we introduce a directional regularizer:
\begin{equation}
\mathcal{L}_{reg}
= \sum_{i=1}^N \sum_{j=1}^M
(1-m_{occ}^{(i,j)}) \, w^{(i)} \,
\bigl(1 - \mathrm{cosSim}\bigl(\mathbf{r}_{j},\, \mathbf{d}_{ij}\bigr)\bigr),
\label{eq:dir_reg}
\end{equation}
\noindent
where $N$ and $M$ denote the numbers of point cameras and camera splats, respectively, $\mathbf{r}_j$ is the camera splat's orientation vector, and $\mathbf{d}_{ij}$ is the unit vector from camera splat $C_j$ to point camera $P_i$. The visibility weight $(1-m_{occ}^{(i,j)})$ is obtained from the self-occlusion mask $I_{occ}^{(i)}$ sampled along the viewing ray from point camera $P_i$ toward camera splat $C_j$, and $w^{(i)}$ is a coverage weight indicating how sufficiently  point camera $P_i$ is observed by the current camera splats.



For each point camera $P_i$, the coverage weight is defined as:
\begin{equation}
w^{(i)} = \frac{
    \sum(\max\left(I_{\text{gt}}^{(i)}- I_{\text{render}}^{(i)},\,0\right)^2 \cdot (1-I_{\text{occ}}^{(i)}))
}{
    \sum{(1-I_\text{occ}^{(i)})}
},
\end{equation}
where the summation goes over the image pixels. Larger $w^{(i)}$ indicates that the current directional sampling around  $P_i$ is insufficient,  thereby encouraging more camera splats toward these under-covered point cameras. 

  
\paragraph*{Boundary regularizer}
We apply a boundary regularizer $\mathcal{L}_{bound}$ to constrain camera splats within a predefined spatial boundary in the camera parameter space. 
This boundary specifies the valid region for camera-splat placement during optimization, and its form differs across capture settings. For inward-facing captures, it keeps splats within a valid distance band around the target object, while for outward-facing captures, it confines them to the free-space region inside the scene.
See the supplementary material for the detailed formulation. 


\paragraph*{Final loss}
We define the final loss function as the sum of the image loss and the directional and boundary regularizers:
\begin{equation}
\mathcal{L}_{total}
= \mathcal{L}_{image}  + \,\mathcal{L}_{reg}  + \,\mathcal{L}_{bound}.
\label{eq:total_loss}
\end{equation}
Since our framework is differentiable, camera splats $C$ are optimized by directly minimizing $L_{total}$.

\begin{figure*}[ht]
    \centering
    \includegraphics[width=\textwidth,trim=0cm 6cm 0cm 0cm, clip]{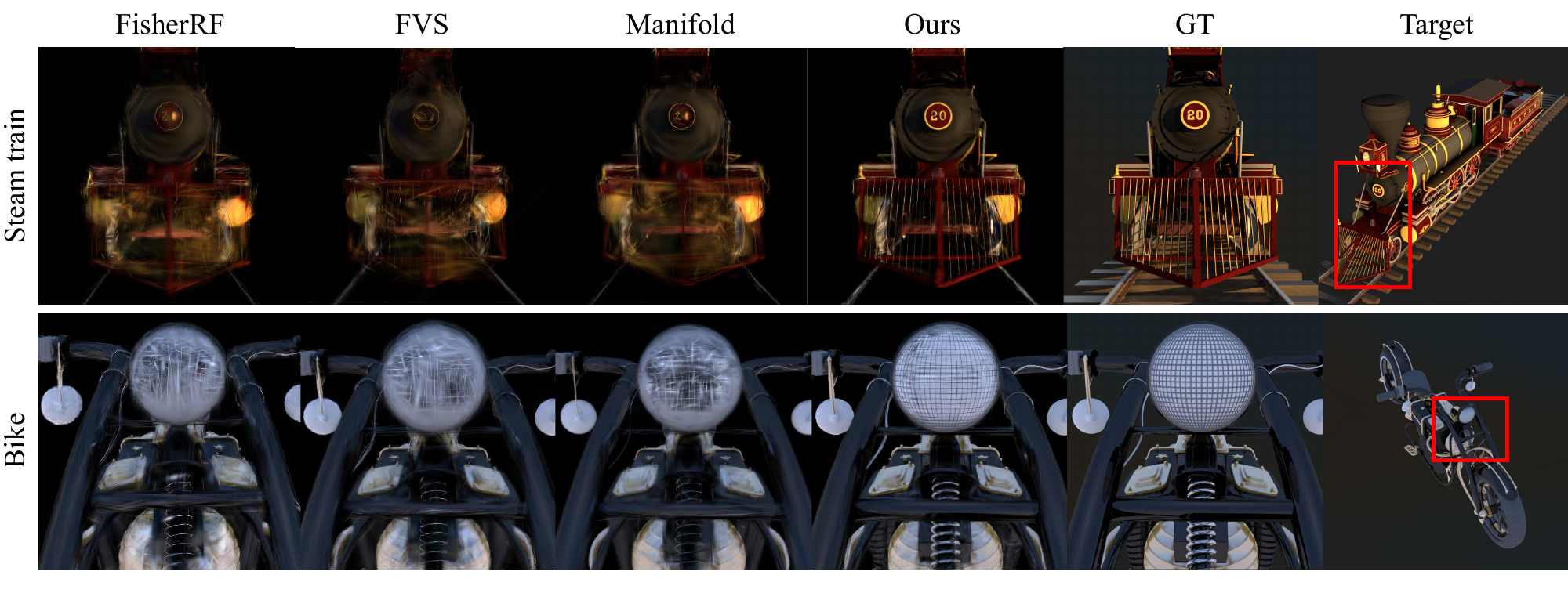}
    \caption{
        Qualitative evaluation on the NSVF datasets.
        The rightmost column indicates the locations of the test-view renderings in the global view. 
        Our view optimization effectively captures detailed geometry and view-dependent appearance.
        }
    \label{fig:nsvf_qual}
    \vspace{-0.2cm}
\end{figure*}

\begin{table*}[t]
    \centering
    \caption{Quantitative comparison on the NSVF test set with 50 and 100 optimized cameras under different test-view sampling settings: Spheres with different radius (half and full radius) and Random. We report PSNR, SSIM, and LPIPS, where higher PSNR/SSIM and lower LPIPS indicate better performance. Best results are highlighted in bold.}
    \label{tbl:nsvf_allcams}
    \Huge
    \renewcommand{\arraystretch}{1.3}
     \resizebox{\textwidth}{!}{
    \begin{tabular}{c ccc ccc ccc ccc ccc ccc}
        \toprule
        & \multicolumn{9}{c}{50 cameras} & \multicolumn{9}{c}{100 cameras} \\
        \cmidrule(lr){2-10}\cmidrule(lr){11-19}
        & \multicolumn{3}{c}{Sphere (half radius)}
        & \multicolumn{3}{c}{Sphere (full radius)}
        & \multicolumn{3}{c}{Random}
        & \multicolumn{3}{c}{Sphere (half radius)}
        & \multicolumn{3}{c}{Sphere (full radius)}
        & \multicolumn{3}{c}{Random} \\
        \cmidrule(lr){2-4}\cmidrule(lr){5-7}\cmidrule(lr){8-10}
        \cmidrule(lr){11-13}\cmidrule(lr){14-16}\cmidrule(lr){17-19}
        & PSNR $\uparrow$ & SSIM $\uparrow$ & LPIPS $\downarrow$
        & PSNR $\uparrow$ & SSIM $\uparrow$ & LPIPS $\downarrow$
        & PSNR $\uparrow$ & SSIM $\uparrow$ & LPIPS $\downarrow$
        & PSNR $\uparrow$ & SSIM $\uparrow$ & LPIPS $\downarrow$
        & PSNR $\uparrow$ & SSIM $\uparrow$ & LPIPS $\downarrow$
        & PSNR $\uparrow$ & SSIM $\uparrow$ & LPIPS $\downarrow$ \\
        \midrule
        
        FisherRF
        & 25.36 &0.819 & 0.180
        & 31.81 & 0.956 & 0.042
        & 28.10 & 0.898 &  0.104
        & 26.78 & 0.856  & 0.157
        & \textbf{33.24} & 0.969&  \textbf{0.035}
        & 29.55 &0.922 &  0.089\\
        
        FVS
        & 25.91 & 0.835 & 0.172
        & \textbf{32.29} & \textbf{0.963} & \textbf{0.038}
        & 28.61 & 0.908 & 0.098
        & 26.70 & 0.857 & 0.159
        & 32.90 & 0.969 & 0.037
        & 29.16 & 0.921 & 0.091  \\

        Manifold
        & 21.77 & 0.743 & 0.220
        & 27.09 & 0.909 & 0.074
        & 23.82 & 0.829 &0.140
        & 22.09 & 0.762 & 0.208
        & 27.99 & 0.920 & 0.069
        & 24.11& 0.840 & 0.132\\
        \midrule
        Ours (Iterative)
        & 24.41 & 0.776 & 0.218
        & 30.71 & 0.950 & 0.050
        & 26.88 & 0.879 & 0.121
        & 25.64 & 0.810 & 0.199
        & 31.75 & 0.960 & 0.046
        & 27.82 & 0.899  & 0.110 \\
        
        Ours (w/o VDS)
        & 26.49 & 0.841 & 0.164
        & 31.25 & 0.959 & 0.041
        & 28.59 & 0.911&  0.094
        & 27.16 & 0.864 & 0.141
        & 32.72 & 0.969 & 0.037
        &29.75 & 0.927 & 0.086  \\
        
        Ours (w/ VDS)
        & \textbf{27.22} & \textbf{0.890} & \textbf{0.150}
        & 31.80 & \textbf{0.963} & \textbf{0.038}
        & \textbf{29.47}& \textbf{0.921} &   \textbf{0.086}
        & \textbf{28.06} & \textbf{0.885} & \textbf{0.135}
        & 32.40 & \textbf{0.970} & \textbf{0.035}
        &  \textbf{30.25} & \textbf{0.935} & \textbf{0.078}  \\
        \bottomrule
    \end{tabular}}
\end{table*}

\subsection{Optimization Process}
Our optimization jointly updates all camera splats in the continuous camera parameter space, as illustrated in \Fig{process}. Given sparse input views, we first reconstruct a coarse proxy geometry as a point cloud using VGGT~\cite{wang2025vggt}. We use 10 views for object-centric scenes and 30 views for scene-centric scenes. For each proxy point, we compute the variance of visible pixel intensities across the input views and map it to a View-Dependency Score (VDS) using the VDSF. The resulting scores are min-max normalized over all proxy points.

We initialize camera splats differently depending on the scene type. For object-centric scenes, splat positions are uniformly sampled on a sphere enclosing the scene, with orientations facing inward toward the object. For scene-centric scenes, splat positions are randomly sampled within a given void-zone bounding box, corresponding to the free space, and their orientations are randomly sampled.

At each iteration, we randomly sample a batch of points from the proxy geometry, using 20 points for object-centric scenes and 200 for scene-centric scenes, and place a point camera at each sampled point. This focuses the optimization on spatial regions near scene surfaces. For each sampled point camera, we construct a ground truth image using precomputed VDS score and render the current camera splats from the same point camera location. We then compute the image loss together with the regularization terms and update all camera splats using gradient-based optimization.

After 1000 optimization iterations, the camera splats converge to the final set of optimized views. We acquire images from these optimized views and use only these images to reconstruct the radiance field. In our experiments, the sparse input views are used solely for the proxy geometry construction and view-dependency estimation, and are not reused for the final reconstruction.

\begin{figure} 
    \centering
    \includegraphics[width=\columnwidth,trim=0cm 9.5cm 0cm 0cm, clip]{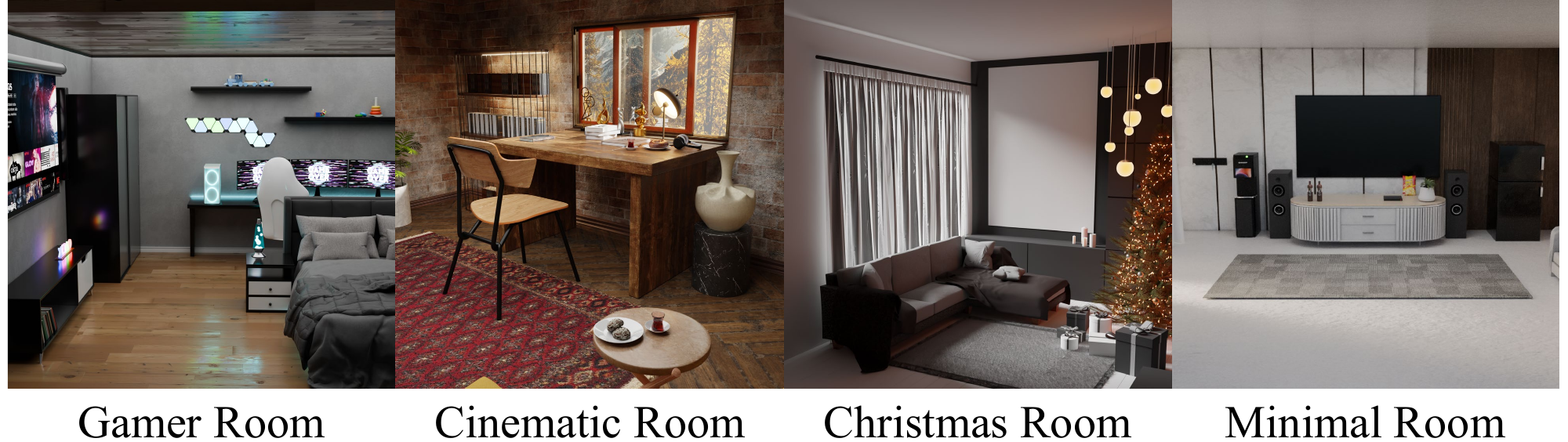}
    \caption{Representative images of scenes from the BlenderKit dataset, covering different levels of scene complexity.
    }
    \label{fig:thumbnail}
    \vspace{-0.2cm}
\end{figure}

\begin{figure*}[ht]
    \centering
    \includegraphics[width=0.95\textwidth,trim=0.5cm 5cm 0cm 0cm, clip]{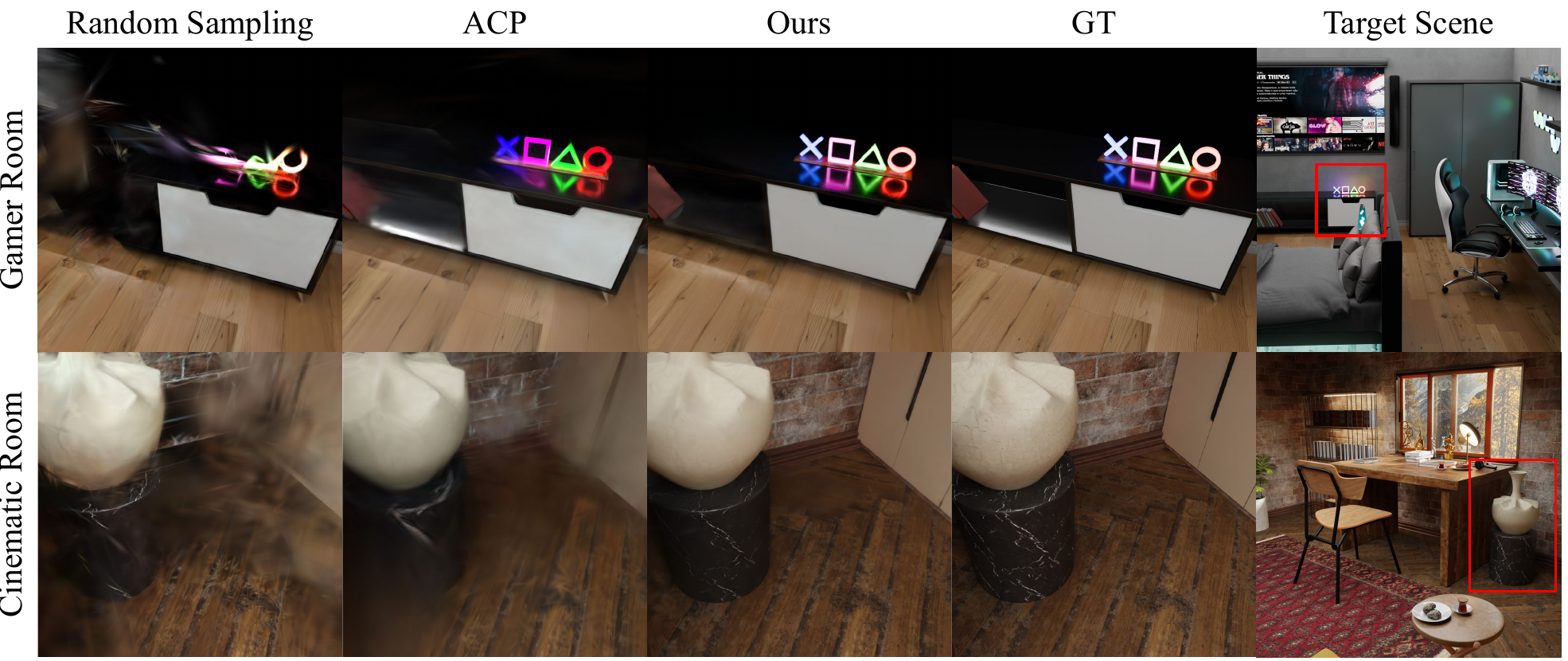}
    \caption{
       Qualitative comparison of different view selection methods on BlenderKit dataset.  The rightmost column shows the target scenes, and the red boxes indicate the regions corresponding to the rendered views.
        }
    \label{fig:quali_blenderkit}
    \vspace{-0.2cm}
\end{figure*}

\begin{table*}     
\vspace{0.2cm}
    \centering
    \caption{
  Quantitative comparison on the BlenderKit test set using 300 optimized views. We report PSNR, SSIM, and LPIPS on four scenes: Gamer Room, Cinematic Room, Christmas Room, and Minimal Room. Higher PSNR/SSIM and lower LPIPS indicate better performance. Best results are highlighted in bold.
    }
    \vspace{-0.2cm}
    \resizebox{1.9\columnwidth}{!}{ 
    \begin{tabular}{ccccccccccccc}
    \toprule
         & \multicolumn{3}{c}{Gamer Room} & \multicolumn{3}{c}{Cinematic Room} & \multicolumn{3}{c}{Christmas Room} & \multicolumn{3}{c}{Minimal Room }  \\
          \cmidrule(lr){2-4}\cmidrule(lr){5-7}\cmidrule(lr){8-10}\cmidrule(lr){11-13}
                    & PSNR \(\uparrow\)  & SSIM \(\uparrow\) & LPIPS \(\downarrow\)  & PSNR \(\uparrow\)  & SSIM \(\uparrow\) & LPIPS \(\downarrow\) & PSNR \(\uparrow\)  & SSIM \(\uparrow\) & LPIPS \(\downarrow\)  & PSNR \(\uparrow\)  & SSIM \(\uparrow\) & LPIPS \(\downarrow\) \\
    \hline
    Random        
        & 23.22 & 0.750 & 0.438 
        & 23.84 & 0.738 & 0.420  
        & 38.27 & 0.940 & 0.115  
        & 31.76 & 0.921 & 0.314 \\
        
    ACP   
        & 23.24 & 0.774 & 0.435 
        & 25.04 & 0.771 & 0.410
        & \textbf{39.87} & 0.948 & 0.110 
        & 32.68 & \textbf{0.935} & 0.298  \\
        
    Ours    
        & \textbf{29.09} & \textbf{0.830} & \textbf{0.364}
        & \textbf{29.79} & \textbf{0.835} & \textbf{0.302}
        & 39.30 & \textbf{0.949} & \textbf{0.105}
        & \textbf{33.22} & 0.933 & \textbf{0.296} \\
        
    \bottomrule
    \end{tabular}
    
    }

    \label{tbl:blenderkit}
\end{table*}


\section{Experiments}
\label{sec:exp} 

\subsection{Setting}
\label{sec:exp_setup}  
\paragraph*{Datasets}
We evaluate our method on three datasets, NSVF~\cite{liu2020neural}, BlenderKit~\cite{blenderkit}, and Tanks and Temples~\cite{Knapitsch2017}, to cover diverse scene settings ranging from synthetic to real-world environments. 

For object-centric evaluation, we use the NSVF dataset, which contains eight diverse synthetic scenes with complex geometry and diverse materials such as glass and metal.
To assess reconstruction quality under diverse viewing positions and directions, we construct three test-view sets. 
We define two sets by sampling cameras on spheres covering the scene with two radii: the full radius, chosen to match the radius of the sphere defining the camera domain used in FVS~\cite{xiao2024nerf} and FisherRF~\cite{jiang2023fisherrf}, and the half radius, set to half that value.
The third set is defined by randomly sampling both camera positions and orientations.
Using different viewing distances allows us to assess reconstruction quality under varying visual appearances and view-dependent effects.
For each set, we sample 200 test views while ensuring that the target object remains visible. 
We evaluate all methods under two camera budgets, 50 and 100 views.

We additionally use four large-scale indoor scenes from BlenderKit~\cite{blenderkit}, namely "Gamer Room", "Cinematic Room",  "Christmas Room" and "Minimal Room" to evaluate the scalability of our method in scene-scale settings that require joint optimization of a large number of cameras. The scenes cover a range of difficulty levels, including cluttered environments with complex textures and illumination, as well as indoor layouts with relatively simple appearance as illustrated in \Fig{thumbnail}.
For each scene, we optimize 300 views and evaluate on 200 test views randomly sampled from the empty space of the scene. 

We also evaluate our method on the Tanks and Temples dataset~\cite{Knapitsch2017} to assess its practicality and generalizability on real-world data. In this setting, the task is formulated as view selection from a discrete image pool, where the goal is to select a subset of 20 or 40 images from a given image pool that yields high-quality radiance field reconstruction. To apply our method in this setting, we first optimize views in the continuous camera space and then select the images in the candidate pool whose camera poses are spatially closest to the optimized views. This allows us to transfer our continuous view optimization framework to a discrete image selection task.

\paragraph*{Comparisons}
We compare our method with prior approaches that represent different optimization strategies. FVS is a geometry-based discrete view selection method and was shown to achieve competitive performance in NeRF Director~\cite{xiao2024nerf}. Manifold Sampling~\cite{lyu2024manifold} is a continuous view optimization method that optimizes views sequentially based on the uncertainty of radiance field. FisherRF~\cite{jiang2023fisherrf} is a Fisher-information-based discrete view selection method that selects informative images from a candidate image pool. ACP~\cite{kopanas2023improving} is a discrete view selection method designed for large-scale indoor scenes. 

\begin{figure*}[ht]
    \centering
    \includegraphics[width=\textwidth,trim=0cm 9cm 0cm 0cm, clip]{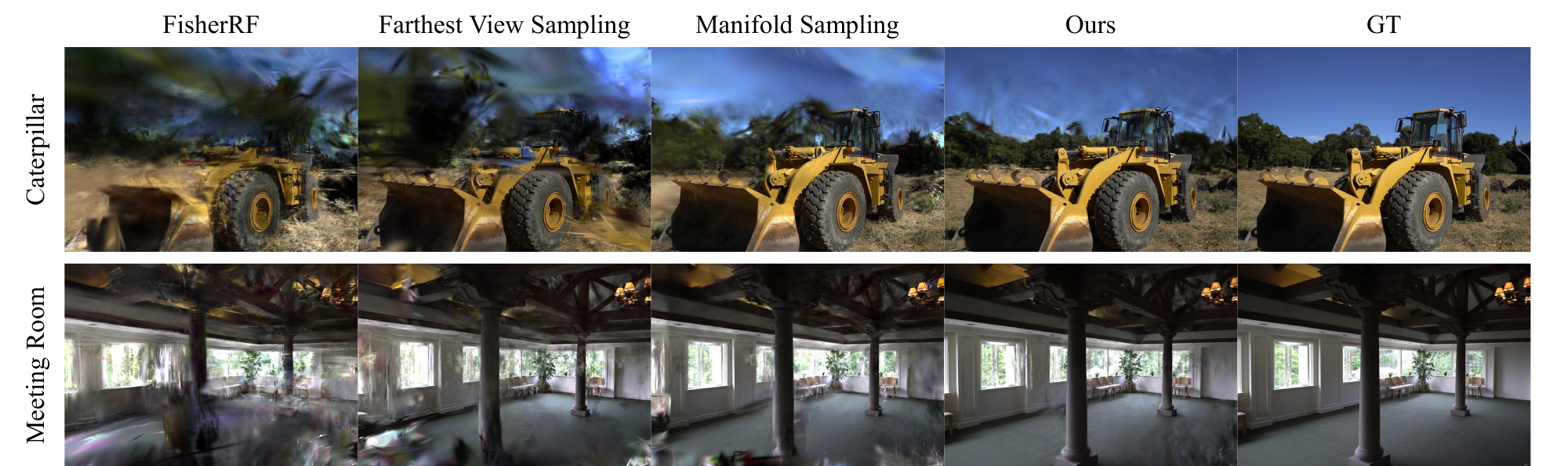}
    \caption{Qualitative comparison of view selection  on the real-world Tanks and Temples dataset.
    }
    \label{fig:tat_quali}
\end{figure*}

We further evaluate three variants of our method: an iterative variant that incrementally adds camera views in a next-best-view manner, optimizing each newly added view while keeping the previously optimized ones fixed (Iterative), a joint optimization variant that uses only proxy geometry without the VDS score (w/o VDS), and the full model, which uses both proxy geometry and the VDS score (w/ VDS).

\paragraph*{Evaluation protocol}
We evaluate all methods by measuring the quality of 3DGS trained on the views selected by each method. For the synthetic datasets, each method first produces a set of optimized views, and the corresponding training images are obtained by rendering the scene at those views. For the real-world dataset, each method selects an optimal subset of views from a given candidate view pool. Corresponding images are then used as inputs for 3DGS reconstruction. The reconstruction quality is evaluated on the test views using PSNR, SSIM, and LPIPS.

\subsection{Continuous View Optimization Evaluation}
\label{sec:exp_quality}
On the NSVF dataset, our method achieves the best overall performance across most evaluation settings (\Tbl{nsvf_allcams}). Relative to FVS and FisherRF, our results highlight the advantage of optimizing camera parameters in a continuous space rather than restricting the solution to a discrete candidate set. 
Relative to Manifold Sampling and our iterative variant, the results further show the benefit of joint optimization over sequential view updates. 
The gains are more pronounced in the half-radius and random settings, where quality gaps in the reconstruction of fine textures and view-dependent appearance are more clearly exposed, as illustrated in \Fig{nsvf_qual}. Among our variants, the full model performs best, indicating that optimizing views according to scene-specific spatial and directional sampling targets leads to more informative view allocation for accurate radiance field reconstruction.

We further evaluate our method on the BlenderKit dataset to examine the effectiveness of jointly optimizing a large number of cameras in scene-scale environments (\Tbl{blenderkit}). Our method outperforms the baselines on most scenes, with larger gains on the more challenging scenes such as "Gamer Room" and "Cinematic Room".  As illustrated in \Fig{quali_blenderkit}, our optimized views better capture the reflective neon sign in "Gamer Room" and the surface texture of the ceramic vase in "Cinematic Room". These results suggest that the proposed framework remains effective as scene complexity and camera budget increase, while allocating views more informatively in regions with complex appearance.

In terms of optimization time on an NVIDIA RTX 4090 GPU, FVS completes view selection in about 1 second, while FisherRF and Manifold Sampling take approximately 13 minutes and 1 hour on the NSVF scenes, respectively.
ACP takes up to 9 hours to optimize 300 views on the BlenderKit scenes. 
In contrast, our method converges in about 1 minute across all datasets, including the BlenderKit scenes with 300 optimized views, demonstrating that our 3DGS-based framework is both efficient and scalable across a wide range of scenes and camera budgets. 


\begin{figure*} 
    \centering
    \includegraphics[width=\textwidth,trim=0.2cm 13cm 0.2cm 0cm, clip]{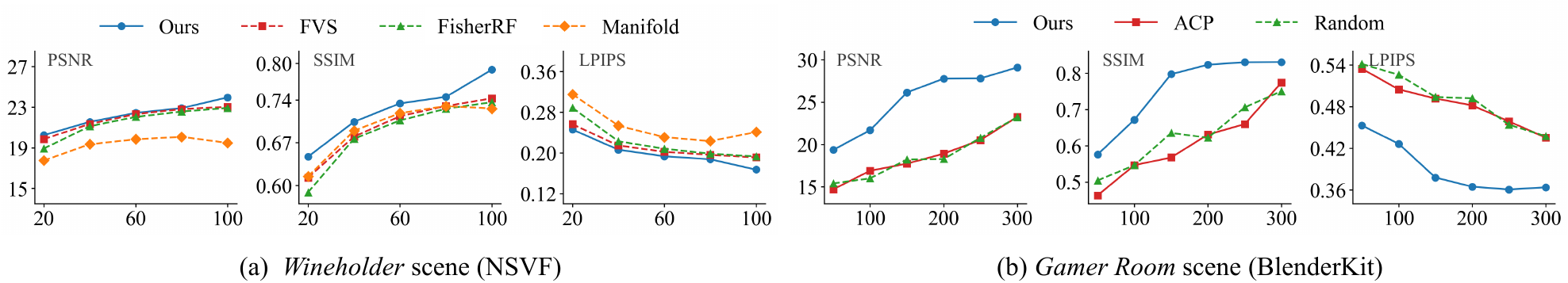}
    \caption{ The graphs show PSNR, SSIM, and LPIPS of radiance fields reconstructed from optimized views under increasing view budgets. The left plots (a) correspond to the Wineholder scene from the NSVF dataset, and the right plots (b) correspond to the Gamer room scene from BlenderKit.
    }
    \label{fig:view_incremental}
    \vspace{-0.2cm}
\end{figure*}

\subsection{Application to Real-world View Selection}
\label{sec:realworld}  
Our framework can also be extended to view selection from real-world image collections. To examine this application, we evaluate our method on the Tanks and Temples dataset~\cite{Knapitsch2017}, where the goal is to select an optimal subset of candidate pool for radiance field reconstruction.  Specifically, we first optimize views in the continuous camera space and then select the input images closest to the optimized views. This provides a practical way to transfer our continuous view optimization framework to a discrete image selection task.

\Fig{tat_quali} and \Tbl{tnt_quanti} show the comparisons of novel view synthesis results at test views. Our method outperforms the baselines under the same camera budgets, indicating that views optimized in continuous space remain informative even when matched to a discrete input pool. These results suggest that our framework generalizes beyond synthetic view optimization and can be effectively applied to real-world scenarios.

\begin{table}     
\vspace{0.2cm}
    \centering
    \caption{
    Quantitative comparison on the test set of the Tanks and Temples dataset for the view selection task, using 20 and 40 optimized cameras. 
    }
    \vspace{-0.2cm}
    \resizebox{\columnwidth}{!}{ 
    \begin{tabular}{ccccccc}
    \toprule
         & \multicolumn{3}{c}{20 cameras} & \multicolumn{3}{c}{40 cameras} \\
          \cmidrule(lr){2-4}\cmidrule(lr){5-7}
                    & PSNR \(\uparrow\)  & SSIM \(\uparrow\) & LPIPS \(\downarrow\)  & PSNR \(\uparrow\)  & SSIM \(\uparrow\) & LPIPS \(\downarrow\)  \\
    \hline
    Manifold         & 12.63  & 0.451  & 0.526  &  14.03 & 0.530  & 0.462 \\
    FVS             & 13.33  & 0.485  & 0.496 & 14.76  & 0.568  & 0.428 \\
    FisherRF         &  12.11 & 0.507  & 0.495  & 14.83 &  0.562 &  0.435 \\ 
    Ours           & \textbf{14.03} &  \textbf{0.511} &  \textbf{0.467} &  \textbf{15.28} & \textbf{0.582}  & \textbf{0.413} \\
    \bottomrule
    \end{tabular} } 

    \label{tbl:tnt_quanti} 
\end{table}

\begin{figure}[t] 
    \centering
    \includegraphics[width=1\linewidth,trim=0.3cm 0.3cm 0.3cm 0.3cm]{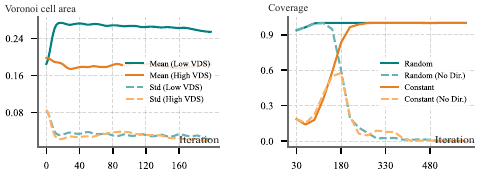}
    \caption{ 
        The left graph analyzes directional sampling density under different VDS targets, using spherical Voronoi cell areas as a proxy for the angular spacing of camera splats.
        The right graph analyzes the effect of the directional regularizer on scene coverage over training iterations under random and identical camera-splat initializations.
        } 
    \label{fig:vds_ablation}
 
\end{figure}

\begin{table*}[t]
    \centering
    \caption{
        Evaluation of robustness to proxy geometry under different input-view configurations. (a) We vary the number of input views used for proxy reconstruction across object-centric NSVF and scene-centric BlenderKit scenes. (b) We fix the number of input views to 10 on the Palace scene from the NSVF dataset, and vary their selection using uniform, random, and biased sampling.
    }
    \label{tbl:ablation_proxy_views}
    \begin{minipage}[t]{0.70\textwidth}
        \centering
        (a) Varying number of input views\\[2pt]
        \resizebox{\linewidth}{!}{
        \begin{tabular}{c cccc cccc cccc}
            \toprule
            & \multicolumn{4}{c}{Palace}
            & \multicolumn{4}{c}{Gamer Room}
            & \multicolumn{4}{c}{Living Room} \\
            \cmidrule(lr){2-5}
            \cmidrule(lr){6-9}
            \cmidrule(lr){10-13}
            & 5 & 10 & 15 & 20
            & 10 & 20 & 30 & 40
            & 10 & 20 & 30 & 40 \\
            \midrule
            PSNR $\uparrow$
            & 28.61 & 29.04 & 29.38 & 29.10
            & 28.85 & 29.05 & 29.09 & 29.23
            & 31.63 & 32.60 & 33.22 & 32.68 \\
            SSIM $\uparrow$
            & 0.702 & 0.718 & 0.729 & 0.716
            & 0.836 & 0.835 & 0.830 & 0.836
            & 0.927 & 0.930 & 0.933 & 0.936 \\
            LPIPS $\downarrow$
            & 0.274 & 0.262 & 0.253 & 0.263
            & 0.359 & 0.362 & 0.364 & 0.359
            & 0.302 & 0.301 & 0.296 & 0.298 \\
            \bottomrule
        \end{tabular}
        }
    \end{minipage}
    \hfill
    \begin{minipage}[t]{0.27\textwidth}
        \centering
        (b) Varying input view selection\\[2pt]
        \resizebox{\linewidth}{!}{
        \begin{tabular}{cccc}
            \toprule
            & Uniform & Random & Biased \\
            \cmidrule(lr){2-2}
            \cmidrule(lr){3-3}
            \cmidrule(lr){4-4}
            & 10 & 10 & 10 \\
            \midrule
            PSNR $\uparrow$
            & 31.01 & 30.92 & 30.52 \\
            SSIM $\uparrow$
            & 0.947 & 0.946 & 0.943 \\
            LPIPS $\downarrow$
            & 0.046 & 0.047 & 0.051 \\
            \bottomrule
        \end{tabular}
        }
    \end{minipage}

\end{table*}

\subsection{Analysis of Camera Splatting}   
\label{sec:ablation} 
\paragraph*{Effect of VDS on directional sampling density}  
We study whether different target view dependency scores (VDS) in \Eq{image_intensity} induce corresponding differences in directional sampling density.
For this analysis, we place point cameras on a sphere and assign each camera either a high or low VDS target.
To quantify directional sampling density, we construct spherical Voronoi diagrams~\cite{caroli2010robust} from the camera splats observed at each point camera.
The area of each Voronoi cell approximates the angular spacing between neighboring camera splats, where smaller cell areas correspond to denser directional sampling.
As shown in the left graph of \Fig{vds_ablation}, point cameras with high VDS targets converge to smaller mean Voronoi cell areas, whereas those with low VDS targets converge to larger mean cell areas.
This indicates that higher VDS targets produce denser directional sampling, whereas lower VDS targets yield sparser directional sampling.
The decreasing standard deviation of the cell areas further indicates that the optimized camera splats become more regularly distributed under both settings.

\paragraph*{Effect of the directional regularizer}
We evaluate the effect of the directional regularizer in \Eq{dir_reg} on robustness to camera-splat initialization, under two settings: (i) random positions oriented toward the scene center and (ii) identical positions and orientations for all camera splats. The second setting is a more challenging extreme case, in which the camera splats provide minimal initial scene coverage and no directional diversity. Scene coverage denotes the fraction of sampled points on the scene surface that are observed by the camera splats.
As shown in the right graph of \Fig{vds_ablation}, the regularizer steers camera splats toward under-covered surface regions, so that both initializations reach nearly full scene coverage even when the second setting starts from a poor initialization. Without the regularizer, the optimization fails to achieve comparable coverage under both initializations.

\subsection{Reconstruction Quality Across View Budgets} 
We evaluate the reconstruction quality of the radiance field as the number of optimized views increases on two scenes with different scales: "Wineholder" from NSVF, an object-centric scene, and "Gamer Room" from BlenderKit, a large-scale indoor scene.
As shown in \Fig{view_incremental}, our method consistently outperforms the baselines across all view budgets on both scenes, indicating that it selects more informative views for reconstruction.
On Wineholder scene, the competing methods tend to saturate as more views are added, whereas our method continues to yield steady gains.
This suggests that, even after the overall scene structure has been largely recovered, our method remains effective at identifying informative views, including those that capture view-dependent appearance effects.
On gamer room, the performance gap is substantial even with few optimized views, showing that our method can effectively select informative views in scenes with complex illumination.

    
        
        
        
        
        

\begin{figure}[ht]
    \centering
    \includegraphics[width=\columnwidth,trim=0cm 0cm 1.75cm 0cm, clip]{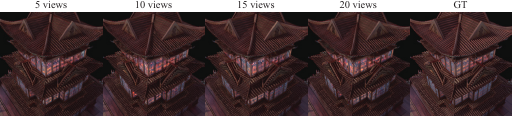}
    \caption{
       Qualitative comparison under different proxy geometry qualities on the palace scene from the NSVF dataset. Each proxy is reconstructed from a different number of input views. The final radiance field reconstruction quality remains visually consistent across the different proxy geometry qualities.
    }
    \label{fig:quali_ablation}
    \vspace{-0.5cm}
\end{figure}

\subsection{Robustness to Proxy Geometry} 
\paragraph*{Varying proxy quality by the number of initial views}
We evaluate the robustness of our method to proxy geometry quality by varying the number of views used for proxy reconstruction. For the object-centric setting, we use the Palace scene from the NSVF dataset, which contains complex geometry, and reconstruct the proxy using 5, 10, 15, and 20 input views.
We further evaluate two scene-centric BlenderKit scenes with different characteristics, Gamer Room and Living Room, using 10, 20, 30, and 40 proxy reconstruction views. For object-centric and scene-centric scenes, we optimize 50 and 300 views, respectively.
As shown in \Tbl{ablation_proxy_views}a and \Fig{quali_ablation}, our method maintains consistent performance across different proxy qualities in both object-centric and scene-centric settings. In the BlenderKit scenes, performance begins to saturate when using 20 or more input views. These results demonstrate robustness to proxy geometry quality across scenes of different complexity and suggest that coarse proxy geometry is sufficient to guide spatial sampling for view optimization.

\paragraph*{Varying proxy quality by initial view selection}
Proxy geometry quality can also vary with the selection of input views. We evaluate three 10-view configurations on the Palace scene: uniform sampling used in our main experiments, random sampling, and biased sampling with nine views in one hemisphere and one in the other. As shown in \Tbl{ablation_proxy_views}b, random sampling performs comparably to uniform sampling, while biased sampling causes only a modest performance degradation. These results further demonstrate robustness to proxy geometry variations induced by different input view selections.

\section{Discussion}

\paragraph*{Rendered intensity as sampling density}
We use the rendered intensity at each point camera as a proxy for the directional sampling density. 
For $k$ camera splats overlapping along a ray, each with opacity $\alpha$, the accumulated intensity is
\begin{equation}
I_k = \sum_{j=0}^{k-1} \alpha (1-\alpha)^j .
\label{eq:accumulated_intensity}
\end{equation}
When both $\alpha$ and $k$ are small, the attenuation term $(1-\alpha)^j$ remains close to one, and the accumulation becomes approximately linear in $k$: $I_k \approx k\alpha$. In our experiments, we use a small opacity, $\alpha=0.1$, and the overlap count $k$ remains small in practice. Thus, higher rendered intensity indicates a higher concentration of camera splats in the corresponding direction.

\paragraph*{Proxy geometry quality}
Our experiments suggest that view optimization is relatively tolerant to moderate proxy inaccuracies. Although fine-scale structures may be reconstructed imperfectly, supervision aggregated from point cameras across the available proxy geometry still guides view optimization. However, if substantial surface regions are missing from the proxy, camera splats may receive insufficient signals to guide them toward those regions, potentially affecting view optimization performance. As a possible preprocessing step, geometry completion using learned or geometric priors could mitigate this issue.

\begin{figure}[t]
    \centering
    \includegraphics[width=0.95\columnwidth,trim=0cm 0cm 0cm 0cm, clip]{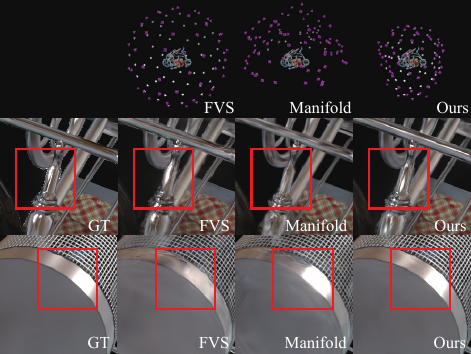}
    \caption{
        Qualitative comparison of optimized view placement and view-dependent appearance recovery.
        The first row illustrates the optimized camera configurations for each method.
        The following rows show novel view synthesis results from radiance fields reconstructed using the optimized views.
        }  
    \label{fig:quali_render} 
\end{figure}

\paragraph*{Direction-aware VDS}
We analyze whether the optimized views are effective at capturing view dependent appearance on the \textit{Wineholder} scene from NSVF dataset (\Fig{quali_render}).
Our method allocates views according to the local View Dependency Score (VDS), placing denser views around highly view-dependent regions such as the metallic wheel and fewer views around less view dependent regions such as the diffuse box.   
While this adaptive allocation improves the recovery of specular highlights, the current VDS simply scales the target intensity uniformly over all viewing directions at each point camera. In addition, since the current VDS is estimated from intensity variance using an empirically fitted function, it may be affected by factors such as lighting changes or exposure variations.
A direction-aware extension of VDS could impose directional weighting over the angular domain, allowing camera splats to concentrate on directions of higher sampling importance. We leave this extension for future work.

\section{Conclusion}
\label{sec:conclusion} 
We introduce Camera Splatting, a novel gradient-based view optimization framework designed for radiance field reconstruction. Departing from discrete view sampling methods, Camera Splatting leverages continuous exploration of the joint view parameter space, effectively capturing complex view-dependent effects and covering the entire scene geometry. 
Benefiting from the computational efficiency of the original 3D Gaussian Splatting, our method achieves rapid evaluation of viewpoint quality, resulting in convergence within one minute, even during simultaneous batch optimization of a large number of views. 
Our framework provides highly informative views, facilitating more efficient and accurate radiance field reconstruction across diverse scenarios and scene scales.

\section*{Acknowledgements}
We thank the anonymous reviewers for the valuable feedback.
This work was supported by NRF grant (RS-2026-25485382, RS-2024-00451947) and IITP grant (RS-2024-00437866) funded by the Korean government (MSIT \& MCST).

\bibliographystyle{eg-alpha-doi} 
\bibliography{egbibsample}       

@String{tog = "ACM TOG"}

@article{smith2018aerial,
  title={Aerial path planning for urban scene reconstruction: A continuous optimization method and benchmark},
  author={Smith, Neil and Moehrle, Nils and Goesele, Michael and Heidrich, Wolfgang},
  year={2018},
  publisher={Association for Computing Machinery (ACM)}
}

@article{fan2016automated,
  title={Automated view and path planning for scalable multi-object 3D scanning},
  author={Fan, Xinyi and Zhang, Linguang and Brown, Benedict and Rusinkiewicz, Szymon},
  journal={ACM Transactions on Graphics (TOG)},
  volume={35},
  number={6},
  pages={1--13},
  year={2016},
  publisher={ACM New York, NY, USA}
}

@article{Knapitsch2017,
    author    = {Arno Knapitsch and Jaesik Park and Qian-Yi Zhou and Vladlen Koltun},
    title     = {Tanks and Temples: Benchmarking Large-Scale Scene Reconstruction},
    journal   = {ACM Transactions on Graphics},
    volume    = {36},
    number    = {4},
    year      = {2017},
}

@article{zhou2020offsite,
  title={Offsite aerial path planning for efficient urban scene reconstruction},
  author={Zhou, Xiaohui and Xie, Ke and Huang, Kai and Liu, Yilin and Zhou, Yang and Gong, Minglun and Huang, Hui},
  journal={ACM Transactions on Graphics (TOG)},
  volume={39},
  number={6},
  pages={1--16},
  year={2020},
  publisher={ACM New York, NY, USA}
}

@article{zhan2022activermap,
  title={Activermap: Radiance field for active mapping and planning},
  author={Zhan, Huangying and Zheng, Jiyang and Xu, Yi and Reid, Ian and Rezatofighi, Hamid},
  journal={arXiv preprint arXiv:2211.12656},
  year={2022}
}

@article{kerbl20233d,
  title={3d gaussian splatting for real-time radiance field rendering.},
  author={Kerbl, Bernhard and Kopanas, Georgios and Leimk{\"u}hler, Thomas and Drettakis, George},
  journal={ACM Trans. Graph.},
  volume={42},
  number={4},
  pages={139--1},
  year={2023}
}

@article{mendoza2020supervised,
  title={Supervised learning of the next-best-view for 3d object reconstruction},
  author={Mendoza, Miguel and Vasquez-Gomez, J Irving and Taud, Hind and Sucar, L Enrique and Reta, Carolina},
  journal={Pattern Recognition Letters},
  volume={133},
  pages={224--231},
  year={2020},
  publisher={Elsevier}
}

@inproceedings{yi2023render,
  title={Where to Render: Studying Renderability for IBR of Large-Scale Scenes},
  author={Yi, Zimu and Xie, Ke and Lyu, Jiahui and Gong, Minglun and Huang, Hui},
  booktitle={2023 IEEE Conference Virtual Reality and 3D User Interfaces (VR)},
  pages={356--366},
  year={2023},
  organization={IEEE}
}

@inproceedings{chen2024gennbv,
  title={GenNBV: Generalizable Next-Best-View Policy for Active 3D Reconstruction},
  author={Chen, Xiao and Li, Quanyi and Wang, Tai and Xue, Tianfan and Pang, Jiangmiao},
  booktitle={Proceedings of the IEEE/CVF Conference on Computer Vision and Pattern Recognition},
  pages={16436--16445},
  year={2024}
}

@inproceedings {kopanas2023improving,
		booktitle = {Vision, Modeling, and Visualization},
		editor = {Guthe, Michael and Grosch, Thorsten},
		title = {{Improving NeRF Quality by Progressive Camera Placement for Free-Viewpoint Navigation}},
		author = {Kopanas, Georgios and Drettakis, George},
		year = {2023},
		publisher = {The Eurographics Association},
		ISBN = {978-3-03868-232-5},
		DOI = {10.2312/vmv.20231222}
}

@inproceedings{xiao2024nerf,
  title={NeRF Director: Revisiting View Selection in Neural Volume Rendering},
  author={Xiao, Wenhui and Cruz, Rodrigo Santa and Ahmedt-Aristizabal, David and Salvado, Olivier and Fookes, Clinton and Lebrat, Leo},
  booktitle={Proceedings of the IEEE/CVF Conference on Computer Vision and Pattern Recognition},
  pages={20742--20751},
  year={2024}
}

@article{lee2022uncertainty,
  title={Uncertainty guided policy for active robotic 3d reconstruction using neural radiance fields},
  author={Lee, Soomin and Chen, Le and Wang, Jiahao and Liniger, Alexander and Kumar, Suryansh and Yu, Fisher},
  journal={IEEE Robotics and Automation Letters},
  volume={7},
  number={4},
  pages={12070--12077},
  year={2022},
  publisher={IEEE}
}

@inproceedings{dunn2009next,
  title={Next Best View Planning for Active Model Improvement.},
  author={Dunn, Enrique and Frahm, Jan-Michael},
  booktitle={BMVC},
  pages={1--11},
  year={2009}
}

@article{li2024frequency,
  title={Frequency-based View Selection in Gaussian Splatting Reconstruction},
  author={Li, Monica MQ and Lajoie, Pierre-Yves and Beltrame, Giovanni},
  journal={arXiv preprint arXiv:2409.16470},
  year={2024}
}

@inproceedings{krainin2011autonomous,
  title={Autonomous generation of complete 3D object models using next best view manipulation planning},
  author={Krainin, Michael and Curless, Brian and Fox, Dieter},
  booktitle={2011 IEEE international conference on robotics and automation},
  pages={5031--5037},
  year={2011},
  organization={IEEE}
}

@article{scott2003view,
  title={View planning for automated three-dimensional object reconstruction and inspection},
  author={Scott, William R and Roth, Gerhard and Rivest, Jean-Fran{\c{c}}ois},
  journal={ACM Computing Surveys (CSUR)},
  volume={35},
  number={1},
  pages={64--96},
  year={2003},
  publisher={ACM New York, NY, USA}
}

@inproceedings{pan2022activenerf,
  title={Activenerf: Learning where to see with uncertainty estimation},
  author={Pan, Xuran and Lai, Zihang and Song, Shiji and Huang, Gao},
  booktitle={European Conference on Computer Vision},
  pages={230--246},
  year={2022},
  organization={Springer}
}

@article{jiang2023fisherrf,
  title={Fisherrf: Active view selection and uncertainty quantification for radiance fields using fisher information},
  author={Jiang, Wen and Lei, Boshu and Daniilidis, Kostas},
  journal={arXiv preprint arXiv:2311.17874},
  year={2023}
}

@inproceedings{jin2023neu,
  title={Neu-nbv: Next best view planning using uncertainty estimation in image-based neural rendering},
  author={Jin, Liren and Chen, Xieyuanli and R{\"u}ckin, Julius and Popovi{\'c}, Marija},
  booktitle={2023 IEEE/RSJ International Conference on Intelligent Robots and Systems (IROS)},
  pages={11305--11312},
  year={2023},
  organization={IEEE}
}

@article{ran2023neurar,
  title={Neurar: Neural uncertainty for autonomous 3d reconstruction with implicit neural representations},
  author={Ran, Yunlong and Zeng, Jing and He, Shibo and Chen, Jiming and Li, Lincheng and Chen, Yingfeng and Lee, Gimhee and Ye, Qi},
  journal={IEEE Robotics and Automation Letters},
  volume={8},
  number={2},
  pages={1125--1132},
  year={2023},
  publisher={IEEE}
}

@inproceedings{lyu2024manifold,
  title={Manifold Sampling for Differentiable Uncertainty in Radiance Fields},
  author={Lyu, Linjie and Tewari, Ayush and Habermann, Marc and Saito, Shunsuke and Zollh{\"o}fer, Michael and Leimk{\"u}hler, Thomas and Theobalt, Christian},
  booktitle={SIGGRAPH Asia 2024 Conference Papers},
  pages={1--11},
  year={2024}
}

@article{savant2024modeling,
  title={Modeling uncertainty for Gaussian Splatting},
  author={Savant, Luca and Valsesia, Diego and Magli, Enrico},
  journal={arXiv preprint arXiv:2403.18476},
  year={2024}
}

@article{liu2020neural,
  title={Neural sparse voxel fields},
  author={Liu, Lingjie and Gu, Jiatao and Zaw Lin, Kyaw and Chua, Tat-Seng and Theobalt, Christian},
  journal={Advances in Neural Information Processing Systems},
  volume={33},
  pages={15651--15663},
  year={2020}
}

@inproceedings{mildenhall2020nerf,
 title={NeRF: Representing Scenes as Neural Radiance Fields for View Synthesis},
 author={Ben Mildenhall and Pratul P. Srinivasan and Matthew Tancik and Jonathan T. Barron and Ravi Ramamoorthi and Ren Ng},
 year={2020},
 booktitle={ECCV},
}

@article{barron2022mipnerf360,
    title={Mip-NeRF 360: Unbounded Anti-Aliased Neural Radiance Fields},
    author={Jonathan T. Barron and Ben Mildenhall and 
            Dor Verbin and Pratul P. Srinivasan and Peter Hedman},
    journal={CVPR},
    year={2022}
}

@inproceedings{kajiya1986rendering,
  title={The rendering equation},
  author={Kajiya, James T},
  booktitle={Proceedings of the 13th annual conference on Computer graphics and interactive techniques},
  pages={143--150},
  year={1986}
}

@inproceedings{fridovich2022plenoxels,
  title={Plenoxels: Radiance fields without neural networks},
  author={Fridovich-Keil, Sara and Yu, Alex and Tancik, Matthew and Chen, Qinhong and Recht, Benjamin and Kanazawa, Angjoo},
  booktitle={Proceedings of the IEEE/CVF conference on computer vision and pattern recognition},
  pages={5501--5510},
  year={2022}
}

@article{kendall2017uncertainties,
  title={What uncertainties do we need in bayesian deep learning for computer vision?},
  author={Kendall, Alex and Gal, Yarin},
  journal={Advances in neural information processing systems},
  volume={30},
  year={2017}
}

@inproceedings{beder2006determining,
  title={Determining an initial image pair for fixing the scale of a 3d reconstruction from an image sequence},
  author={Beder, Christian and Steffen, Richard},
  booktitle={Joint Pattern Recognition Symposium},
  pages={657--666},
  year={2006},
  organization={Springer}
}

@inproceedings{sunderhauf2023density,
  title={Density-aware nerf ensembles: Quantifying predictive uncertainty in neural radiance fields},
  author={S{\"u}nderhauf, Niko and Abou-Chakra, Jad and Miller, Dimity},
  booktitle={2023 IEEE International Conference on Robotics and Automation (ICRA)},
  pages={9370--9376},
  year={2023},
  organization={IEEE}
}

@inproceedings{roberts2017submodular,
    author       = {Mike Roberts AND Debadeepta Dey AND Anh Truong AND Sudipta Sinha AND
                    Shital Shah AND Ashish Kapoor AND Pat Hanrahan AND Neel Joshi},
    title     = {Submodular Trajectory Optimization for Aerial 3D Scanning},
    booktitle = {International Conference on Computer Vision (ICCV) 2017},
    year      = {2017}
}

@inproceedings{caroli2010robust,
  title={Robust and efficient Delaunay triangulations of points on or close to a sphere},
  author={Caroli, Manuel and de Castro, Pedro MM and Loriot, S{\'e}bastien and Rouiller, Olivier and Teillaud, Monique and Wormser, Camille},
  booktitle={Experimental Algorithms: 9th International Symposium, SEA 2010, Ischia Island, Naples, Italy, May 20-22, 2010. Proceedings 9},
  pages={462--473},
  year={2010},
  organization={Springer}
}

@Webpage{blenderkit,
  author = {BlenderKit},
  url = {https://github.com/BlenderKit/BlenderKit},
    year = 2021,
  lastchecked = {23 March 2025}
}

@inproceedings{wang2025vggt,
  title={VGGT: Visual Geometry Grounded Transformer},
  author={Wang, Jianyuan and Chen, Minghao and Karaev, Nikita and Vedaldi, Andrea and Rupprecht, Christian and Novotny, David},
  booktitle={Proceedings of the IEEE/CVF Conference on Computer Vision and Pattern Recognition},
  year={2025}
}

@InProceedings{Wang_2024_CVPR_dust3r,
    author    = {Wang, Shuzhe and Leroy, Vincent and Cabon, Yohann and Chidlovskii, Boris and Revaud, Jerome},
    title     = {DUSt3R: Geometric 3D Vision Made Easy},
    booktitle = {Proceedings of the IEEE/CVF Conference on Computer Vision and Pattern Recognition (CVPR)},
    month     = {June},
    year      = {2024},
    pages     = {20697-20709}
}



\end{document}